\documentclass[journal]{IEEEtran}
\ifCLASSINFOpdf
\else
\fi
\usepackage{mathrsfs}
\usepackage{amssymb}
\usepackage{algorithm}
\usepackage{amsmath}
\usepackage{amsthm}
\usepackage{algpseudocode}
\usepackage{array}
\usepackage{booktabs} 
\usepackage{bm}
\usepackage{color}
\usepackage{colortbl}
\usepackage{enumitem}
\usepackage{epsfig}
\usepackage{float}
\usepackage{graphicx}
\usepackage{lineno}
\usepackage{setspace}
\usepackage{indentfirst}
\usepackage{multirow}
\usepackage{algorithm}
\usepackage{microtype}
\usepackage{graphicx}
\usepackage{subfigure}
\usepackage{booktabs} 
\usepackage{amssymb}
\usepackage{enumerate}
\usepackage{mathrsfs}
\usepackage{graphicx}
\usepackage{float}
\usepackage{algorithm}
\usepackage{amsmath}
\usepackage{amsthm}
\usepackage{amssymb}
\usepackage{lineno}
\usepackage{setspace}
\usepackage{indentfirst}
\usepackage{epsfig}
\usepackage{multirow}
\usepackage{bm}
\usepackage{makecell} 
\usepackage{color}
\usepackage{colortbl}
\usepackage{lineno}
\usepackage{amssymb}
\usepackage{mathrsfs}
\usepackage{graphicx}
\usepackage{float}
\usepackage{algorithm}
\usepackage{amsthm}
\usepackage{amssymb}
\usepackage{setspace}
\usepackage{indentfirst}
\usepackage{epsfig}
\usepackage{multirow}
\usepackage{bm}
\usepackage[ragged]{footmisc}
\usepackage{scrextend}

\usepackage{amsmath}

\begin{document}
%
\title{Adaptive Nearest Neighbor : A General Framework for Distance Metric Learning}
%
%
%
%
\author{Kun Song 
%
\IEEEcompsocitemizethanks{
\IEEEcompsocthanksitem Kun Song and Junwei Han are with School of Automation, Northwestern Polytechnical University, China.\protect\\
E-mail: \{junweihan2010, songkun123000\}@gmail.com
\IEEEcompsocthanksitem
}
}
\maketitle

\begin{abstract}
$K$-NN classifier is one of the most famous classification algorithms, whose performance is crucially dependent on the distance metric. When we consider the distance metric as a parameter of $K$-NN, learning an appropriate distance metric for $K$-NN can be seen as minimizing the empirical risk of $K$-NN. In this paper, we design a new type of continuous decision function of the $K$-NN classification rule which can be used to construct the continuous empirical risk function of $K$-NN. By minimizing this continuous empirical risk function, we obtain a novel distance metric learning algorithm named as adaptive nearest neighbor (ANN). We have proved that the current algorithms such as the large margin nearest neighbor (LMNN), neighbourhood components analysis (NCA) and the pairwise constraint methods are special cases of the proposed ANN by setting the parameter different values. Compared with the LMNN, NCA, and pairwise constraint methods, our method has a broader searching space which may contain better solutions. At last, extensive experiments on various data sets are conducted to demonstrate the effectiveness and efficiency of the proposed method.
\end{abstract}
\section{Introduction}
\indent Classification is a fundamental task in the field of machine learning. Although the superior performance on classification tasks is obtained by deep learning classifiers, it is still meaningful to study the conventional classifiers. One side of the reasons is the knowledge of traditional classifiers would give some useful inspiration to the design of deep learning classifiers. For example, the cross-entropy loss widely used in deep learning classifiers comes from the traditional softmax regression \cite{zhang2018generalized, kawaguchi2016deep,ustinova2017multi}. The other side is that traditional classifiers are still valuable on the task with small data set \cite{dong2019learning,ye2019learning}.\\
\indent $K$-NN classifier is one of the most appealing classifiers due to its effectiveness, easy implementation, and the ability to handle the non-linear problem. However, the performance of $K$-NN classifier is crucially dependent on the distance metric used to measure the similarity between two samples. Hence, it is very beneficial to find one way to find a suitable distance metric for $K$-NN. Distance metric learning is one class of learning paradigms which find distance metric for specifical applications. In past decades, plenty of distance metric learning algorithms have been proposed. The survey of distance metric learning is referred to \cite{wang2015survey,kulis2013metric,yang2006distance,yang2007overview,wang2018survey}. The excellent examples include LMNN \cite{weinberger2009distance}, ITML \cite{davis2007information}, NCA \cite{Goldberger2004Neighbourhood}, GMML \cite{zadeh2016geometric}, RVML \cite{perrot2015regressive}, and so on. Besides those conventional algorithms, with the boom of deep learning, many deep metric learning algorithms have been proposed，such as faceNet \cite{Schroff2015FaceNet}, energy-based neural networks \cite{chopra2005learning}, lifted structured feature embedding \cite{oh2016deep}, to name a few. Those deep metric learning algorithms benefit a lot from the knowledge of the traditional metric learning methods.\\
\indent Although most of the methods mentioned above can provide metric to $K$-NN classifier, the algorithms designed specifically to improve $K$-NN have attract more attention in the classification applications, due to their better performance. They can mainly be classified into two groups.\\
\indent The first class of methods is triplet-based methods whose performance is dependent on the prior information used to construct the triplets. The representative examples include large margin nearest neighbor (LMNN), faceNet, and energy-based neural network (EBNN) and their variants \cite{perrot2015regressive1,Di2017Large,Torresani2006Large,song2017parameter}. When those methods are applied in practice, they may suffer from following three shortcomings. \\
\indent The first is that the triplet constraints are too strict for $K$-NN classification rule. The goal of triplet constraints is to keep the local neighborhood of each inquiry sample being pure, i.e, no imposters located in the local neighborhood \cite{weinberger2009distance}. Indeed, when all of the imposters are out of the local neighborhood of the inquiry sample, the inquiry sample is classified correctly and the training error of $K$-NN is reduced. However, this is not a prerequisite for correctly classifying the inquiry sample by using the $K$-NN rule. That is because the $K$-NN rule outputs the classification result according to the majority of labels of the test samples' $K$ nearest neighbors. Even though some imposters are in the local neighborhood, so long as they are not the majority, the $K$-NN rule would output the correct result. To this extent, the constraints of triplet-based methods are too strict, which would narrow the searching space of the learned metric matrix.\\
 \indent Besides, how to provide the prior information is another important problem. The prior information is used to identify the hard region and reduce the number of imposters in the hard region. The so-called hard region is a small area of the local neighborhood in which the imposters could never be pushed out, no matter the value of the metric is. So, if the imposters are located in the hard region, they would produce many harmful triplets which would hurt the training of the model, sometimes they would make the training completely fail \cite{Schroff2015FaceNet,Liang2018Adaptive,perrot2015regressive1}. So how to provide the prior information is very important to the triplet-based model. At last, the number of triplet constraints is about $O(KN^2)$ where $K$'s maximal value can reach $N$. It is too huge for many real applications since they require plenty of computation and storage sources. Therefore, how to reduce the number of triplet constraints is another problem.\\
\indent The second class of methods does not require any prior information to depict the local neighborhood of samples. The representative example is NCA which is designed to minimize the expected one-of-leave training error of the statistical nearest neighbor (SNN) classification rule. In SNN, the classification contribution of the samples with the same label of the inquiry sample and that of the samples with different labels from the inquiry are modeled by the same distribution, i.e., the softmax function. However, the softmax function model tends to give large weights to the nearest samples, and give small weights to the rest of ones. Statistically, the samples with different labels from the inquiry sample are farther than the samples with the same label of inquiry sample. So the softmax function model would ignore many dissimilar samples in the training procedure. This would make NCA lost many discriminant information which may benefit to the classification.\\
\indent Essentially, when we consider the distance metric as a parameter of $K$-NN classifier, a distance metric learning method can be seen as the one to find the parameter minimizing the empirical risk of $K$-NN classifier. From this perspective, the shortcomings of the methods mentioned above are caused by the gap between their loss function and the empirical risk function of $K$-NN classifier. This inspires us to design a distance metric learning algorithm to directly minimizing the empirical risk of $K$-NN classifier. However, since the procedure of classification involves the finding of nearest neighbors, the decision function of the $K$-NN rule is non-continuous to the distance metric. Therefore, it is hard to minimize the empirical risk of the $K$-NN classification rule like the other classifiers such as support vector machines (SVMs), linear regression (LR) and softmax regression (SR) \cite{friedman1997bias}.\\
\indent In this paper, we design an interesting continuous function that can compute the average value of the $K$ smallest numbers and the average value of the $K$ largest numbers in a large series of numbers. This smart continuous function can help us design a continuous decision function of $K$-NN rule. By utilizing the new continuous decision function of $K$-NN, we formulate a novel distance metric learning model named adaptive nearest neighbor (ANN) which directly minimizes the empirical risk of $K$-NN.\\
\indent The contributions of the paper are listed as follows:
\begin{itemize}
 \item We have proved that the LMNN, neighbourhood components analysis (NCA) \cite{Goldberger2004Neighbourhood} and the naive form of pairwise constraint metric learning algorithm (PML) \cite{hu2017sharable,ye2019learning} are special cases of the proposed ANN. Therefore, our work has built up a connection between the convex distance metric learning model LMNN, the convex pairwise-based method and the non-convex metric learning model NCA.
  \item Since the objective function of ANN can be seen as the empirical risk of the $K$-NN classification rule, the searching space of the ANN is more accurate than that of LMNN, NCA, and pairwise constraint methods. This implies that the proposed method may achieve better performance.
  \item In the proposed ANN, there are only $N(N-1)$ distance computations involved in the calculation of the gradient of the objective function, the amount of distance computations is comparable to the pairwise constraint metric learning algorithms. Therefore, the running speed of the proposed method is faster than that of LMNN.
  \item We evaluate the proposed method by conducting extensive experiments of classification tasks on several data sets. The promising results have demonstrated the superiority of the proposed method.
\end{itemize}
\begin{figure}[t]
 \centering
  \includegraphics[width=0.6\linewidth]{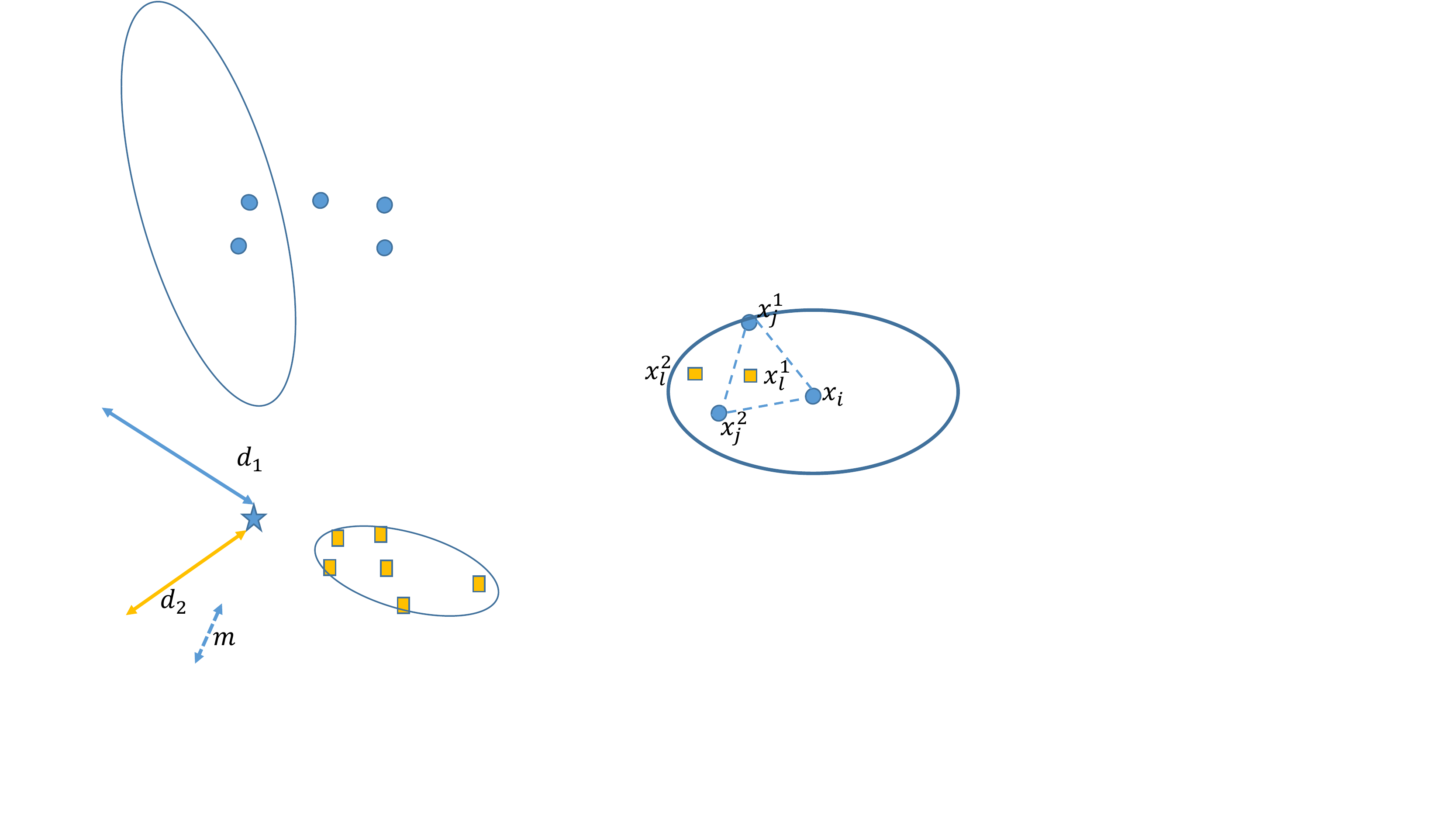}
  \caption{The illustration of the harmful triplets. The blue circle vertexes $\textbf{x}_j^1$ and $\textbf{x}_j^2$ are two target neighbors of $\textbf{x}_i$, and yellow squares $\textbf{x}_l^1$ and $\textbf{x}_l^2$ are imposters. The ellipse is local neighborhood is determined by $\textbf{x}_j^1$ and $\textbf{x}_j^2$. There is no metric $\textbf{M}$ can let the $\textbf{x}_l^1$ out of the local neighborhood. Therefore, the region in the triangle is the hard region.}
  \label{fig_neb}
\end{figure}
\section{Related Works}
\subsection{Decision Function of $K$-NN Classifier }
\indent In this section, we introduce a new type of decision function of $K$-NN classifier proposed recently by \cite{dong2019learning}. This new type of $K$-NN classifier is started from the two-class case, i.e., to judge whether a sample belongs to a specially-pointed class. The bi-class decision function is presented as follows.
  \begin{equation}\label{nn_funK}
 h(\textbf{x}) = \frac{1}{K}\sum_{k=1}^Kd^{[k]}(\textbf{x},\mathcal{C}_c) - \frac{1}{K}\sum_{k=1}^Kd^{[k]}(\textbf{x},(\bar{\mathcal{C}_c}))
 \end{equation}
 where $\textbf{x}$ is the test sample, $\mathcal{C}_c$ represents the collection of samples in the $c$-th class, and $\bar{\mathcal{C}_c}$ represents the collection of samples not in the $c$-th class. $ d^{[k]}(\textbf{x},\mathcal{C})$ is the $k$-th smallest element in set $\{d(\textbf{x},\textbf{x}')|\textbf{x}' \in \mathcal{C}\}$, where $d(\textbf{x},\textbf{x}')$ is the distance function between sample $\textbf{x}$ and $\textbf{x}'$. Considering $h(\textbf{x})$, when $h(\textbf{x}) <0$, $\textbf{x}$ is classified to the class $c$, otherwise, $\textbf{x}$ is not classified to class $c$.\\
\indent For the multi-class tasks, the decision function of $K$-NN classifier can be obtained by extending Eq.(\ref{nn_funK}) according to the `one-vs-rest' strategy \cite{hong2008probabilistic}. The formulation is presented as follows:
 \begin{equation}\label{Knn_fun}
 y = \text{argmin}_c\frac{1}{K}\sum_{k=1}^Kd^{[k]}(\textbf{x},\mathcal{C}_c)
 \end{equation}
 where $\textbf{x}$ is the test sample and $y$ is the predicted label of $\textbf{x}$.
\subsection{Large Margin Nearest Neighbor (LMNN)}
  Let $\mathcal{X}=\{(\textbf{x}_i,y_i)\}^N_{i=1}$ denote the training set consisting of $N$ labeled examples in which $\textbf{x}_i\in \mathbb{R}^{d\times 1}$ and its corresponding class label $y_i\in \{1,2,\cdots,C\}$, where $C$ is the class number. Given a metric matrix $\textbf{M}\in \mathbb{R}^{d\times d}$, the square distance is determined as
\begin{equation}
\label{eq2}
\begin{split}
d_\textbf{M}(\textbf{x}_i,\textbf{x}_j) &={({\textbf{x}}_i - {\textbf{x}}_j)^T\textbf{M}({\textbf{x}}_i - {\textbf{x}}_j)}\\
\end{split}
\end{equation}
where $\textbf{M} \in \mathbb{R}^{d\times d}\succeq 0$. \\
\indent Now, we describe how to construct the triplet constraints. For the $i$-th sample $(\textbf{x}_i,y_i)$ in $\mathcal{X}$, two groups of samples are selected from the rest samples in $\mathcal{X}-\{x_i\}$. The first group is the target neighbors denoted as $\mathcal{P}_i$ which consists of samples with label $y_i$. The second group is imposters denoted as $\mathcal{I}_i$ which consists of samples with different labels from $y_i$. The triplet set associated with $\textbf{x}_i$ is denoted as $\mathcal{T}_i=\{(\textbf{x}_i,\textbf{x}_j,\textbf{x}_l)|\textbf{x}_j \in \mathcal{P}_i, \textbf{x}_l \in \mathcal{I}_i\}$. Each triplet in $\mathcal{T}_i$ represent a constraint $d_\textbf{M}(\textbf{x}_i,\textbf{x}_j)<d_\textbf{M}(\textbf{x}_i,\textbf{x}_l)$. Those triplet constraints force that all of the imposters in $\mathcal{I}_i$ should be out of the local neighborhood $\mathcal{A}_i = \cup_{j\in \mathcal{P}_i}\{\textbf{x}|d_\textbf{M}(\textbf{x},\textbf{x}_i) < d_\textbf{M}(\textbf{x}_i,\textbf{x}_j)\}$. The model of LMNN is formulated as:
\begin{equation}
\label{eq3}
\begin{split}
\min_{\textbf{M} \succeq 0}\sum_{i=1}^N\sum_{j \in \mathcal{P}_i}\sum_{l \in \mathcal{I}_i}\ell(d_\textbf{M}(\textbf{x}_i,\textbf{x}_j)\!\!-\!\!d_\textbf{M}(\textbf{x}_i,\textbf{x}_l)) + \lambda \Omega(\textbf{M})
\end{split}
\end{equation}
where $\ell(x) $ is the hinge loss, i.e., $max\{1-x,0\}$, and $\Omega(\textbf{M}) = \sum_{i=1}^N\sum_{j \in \mathcal{P}_i} d_\textbf{M}(\textbf{x}_i,\textbf{x}_j)$ can be seen as the regularization term to avoid the elements of $\textbf{M}$ being too large, and $\lambda$ is used to balance the two terms in the objective. The optimization problem in Eq.(\ref{eq3}) is often solved by the gradient descent methods.\\
\indent Since the LMNN is the seminal work of triplet-based distance metric learning, we take the LMNN as the example to demonstrate the shortages of triplet-based methods. The reason why the triplet constraints are too restrict to $K$-NN classification rule is obvious. So we only introduce why the prior information is important to the triplet-based methods.\\
\indent As seen from the Fig. {\ref{fig_neb}}, the inquiry sample $\textbf{x}_i$ and its two target neighbors $\textbf{x}_j^1$ and $\textbf{x}_j^2$ form a trigon $\triangle\widehat{\textbf{x}_i\textbf{x}^1_j\textbf{x}_j^2}$ in the feature space. The ellipse represents the local neighborhood $\mathcal{A}_i$ determined by target neighbors  $\textbf{x}_j^1$ and $\textbf{x}_j^2$. Obviously, the trigon $\triangle\widehat{\textbf{x}_i\textbf{x}^1_j\textbf{x}_j^2}$ is in the local neighborhood $\mathcal{A}_i$. Therefore, when the imposter $\textbf{x}_l^1 \in \mathcal{I}_i$ located in the trigon, no matter what the value of metric $\textbf{M}$ is attended, $\textbf{x}^1_l$ could not be pulled out of the local neighborhood. We call $\triangle\widehat{\textbf{x}_i\textbf{x}^1_j\textbf{x}_j^2}$ as the hard region, and imposters located in the hard region as hard imposters. Therefore, the triplet constraints consisting of hard imposters do not provide any useful information and narrow the searching space. There are two ways to remove those harmful triplet constraints. The first is to select the target neighbors as closer as possible, which would narrow the area of the hard region. The second way is to remove the imposters located in the hard area. The first way is widely used in the LMNN and its variants \cite{Di2017Large,song2017parameter,Torresani2006Large,weinberger2005distance}, while the second way is widely used in the triplet-based deep metric learning \cite{Schroff2015FaceNet,Yin2016Fine,cheng2016person}. However, both ways require the prior information to measure the similarity of samples.\\
\subsection{Neighborhood Component Analysis}
\indent Neighbourhood Components Analysis (NCA) \cite{Goldberger2004Neighbourhood} minimizes the expected leave-one-out training error of the statistics nearest neighbor (SNN) which is modeled by the softmax function. Specially, for an inquiry sample $\textbf{x}_i$, the sample $\textbf{x}_j$ gives the probability to let $\textbf{x}_i$ in the $y_j$-th class  is measured by
\begin{equation}\label{NCA_m}
p_{ij} =\left\{
                 \begin{array}{c}
                   \frac{exp(-d_\textbf{M}(\textbf{x}_i,\textbf{x}_j))}{\sum_{i \neq k}exp(-d_\textbf{M}(\textbf{x}_i,\textbf{x}_k))}, i \neq j\\
                  \quad\quad\quad 0\quad\quad\quad\quad\quad, i = j\\
                 \end{array}
               \right.
\end{equation}
Thus, the probability that $\textbf{x}_i$ is classified into the $c$-th class is computed as
\begin{equation}\label{NCA_pp1}
  p_i = \sum_{j\in \mathcal{C}_c} p_{ij}
\end{equation}
Therefore, the optimization problem of NCA is presented as follows.
\begin{equation}\label{NCA_2}
J(\textbf{M}) = \max_{\textbf{M}\succeq 0 }\sum_{i=1}^N\sum_{j \in \mathcal{C}_{y_i}} p_{ij}
\end{equation}
where $\mathcal{C}_{y_i}$ is the set of samples in the class $y_i$ which is the label of sample $\textbf{x}_i$.\\
\indent Obviously, NCA has a broader searching space compared with LMNN, since it directly minimizes the training error of the $1$-NN classification rule. However, the performance of NCA highly relies on the match between the data distribution and the probability distribution described by the softmax function. For example, the samples in different classes may distribute differently. Therefore, it is not accurate to describe the contributions of samples in different classes to classification in one probability function, i.e., the softmax function. The softmax function tends to give a small portion of samples very large probability and other samples very small probability, therefore, $p_i$ in Eq.(\ref{NCA_pp1}) may be very large while $1-p_i$ would be small. In this situation, the action of maximizing $p_i$ would not consider discriminant information. This would harm the performance of NCA.\\
\section{Adoptive Nearest Neighbor for Distance Metric Learning}
In this section, we first introduce the empirical loss of the $K$-NN rule. Then, we propose the formulation of the adoptive nearest neighbor (ANN) by finding a way to make the non-continuous empirical loss continuous. Thirdly, a method to solve the ANN model is introduced. At last, we discuss the connections between ANN and other distance metric learning algorithms.\\
 \begin{algorithm}[tb]
   \caption{Gradient Descent Method for ANN}
   \label{algexample}
\begin{algorithmic}
   \State {\bfseries Input:} Data set $\{(\textbf{x}_i,y_i)\}_{i=1}^N$, parameters $\alpha$, $\lambda$, max iteration number $s_m$.
   \State {\bfseries Output:} Projection matrix $\textbf{M}$.
   \State Initialize $\textbf{M}$ by $\textbf{M}_0$, step-size $\eta$ by $\eta_0$.
   \For{$i=1$ to $N$}
   \State Construct $\mathcal{S}_i$ and $\mathcal{D}_i$.
   \EndFor
   \State Compute objective function $J_0=J_1(\textbf{M}_0)$ by Eq.(\ref{jj_obj})
   \For{$j=1$ to $s_m$}
   \State Compute gradient $\nabla \textbf{M}= \frac{\partial J(\textbf{M})}{\partial \textbf{M}}|_{\textbf{M}=\textbf{M}_{j-1}}$ by Eq.(\ref{dml_ipr});
   \State $\hat{\textbf{M}}= \varphi (\textbf{M}_{j-1} - \eta\nabla \textbf{M})$
   \State  Compute objective function $J(\hat{\textbf{M}})$
   \If{$J(\hat{\textbf{M}})< J_{i-1}$}
   \State $\textbf{M}_{j} =\hat{\textbf{M}}$, $J_j=J(\hat{\textbf{M}})$, $\eta = 1.05\eta$
   \Else
   \State $\eta = 0.5\eta$
   \EndIf
   \EndFor
\end{algorithmic}
\end{algorithm}
 \subsection{The Empirical Risk of $K$-NN classifier}
\indent  Given a training data set $\mathcal{X} =\{({\textbf{x}}_i,y_i)\}_{i=1}^N$, where ${\textbf{x}}_i \in \mathbb{R}^d$, its corresponding label $y_i \in \{1,\cdots,C\}$. For the $i$-th inquiry ${\textbf{x}}_i$, we use $\mathcal{S}_i$ to denote similarity set consisting of samples with same label to $y_i$, and use symbol $\mathcal{D}_i$ to denote set of samples with different labels from $y_i$.\\
\indent Without loss of generality, we use symbol $d_\textbf{M}^{[k]}(\textbf{x}_i,{\mathcal{S}_i})$ to represent the $k$-th smallest element in the distance set $\{d_\textbf{M}(\textbf{x}_i,\textbf{x}_j)|, j \in \mathcal{S}_i\}$ under the metric $\textbf{M}$. Similarly, we use $d_\textbf{M}^{[k]}(\textbf{x},{\mathcal{D}_i})$ to denote the $k$-th smallest element in $\{d_\textbf{M}(\textbf{x}_i,\textbf{x}_l)|, l \in \mathcal{D}_i\}$. According to the decision function in Eq.(\ref{nn_funK}), if the inquiry $\textbf{x}_i$ is classified correctly, the following inequality holds:
 \begin{equation}\label{imtrco1}
\frac{1}{K}\sum_{k=1}^Kd_\textbf{M}^{[k]}\left(\textbf{x}_i,{\mathcal{S}_i}\right) < \frac{1}{K}\sum_{k=1}^Kd_\textbf{M}^{[k]}\left(\textbf{x}_i,{\mathcal{D}_i}\right)
 \end{equation}
 For convenience, we define two functions as follows
 \begin{equation}\label{kneighborf}
 \begin{split}
d^s_{x_i}(\textbf{M},K) &= \frac{1}{K}\sum_{k=1}^Kd_\textbf{M}^{[k]}\left(\textbf{x}_i,{\mathcal{S}_i}\right)\\
d^d_{x_i}(\textbf{M},K) & =\frac{1}{K}\sum_{k=1}^Kd_\textbf{M}^{[k]}\left(\textbf{x}_i,{\mathcal{D}_i}\right)
 \end{split}
 \end{equation}
\indent Similar to traditional classifiers, we can formulate the empirical risk function of $K$-NN classifier as the objective of following model:\\
   \begin{equation}\label{imtco2}
  \begin{split}
  &\min_{\textbf{M}\succeq 0} \sum_{i=1}^N \ell\left( d^s_{x_i}(\textbf{M},K)- d^d_{x_i}(\textbf{M},K)\right) + \lambda\Omega(\textbf{M})
  \end{split}
 \end{equation}
 where $\ell\left(x\right)$ is a loss function which penalizes large $x$, and $\lambda$ is the coefficient of the regularization term $\Omega(\textbf{M})$. \\
 \indent \textbf{Remark 1:} Since the model in Eq.(\ref{imtco2}) involves in the selection of the $K$ nearest neighbors, the corresponding objective function is non-continuous. Thus, the proposed optimization problem in Eq.(\ref{imtco2}) can not be solved by the gradient decent method. The \textbf{Lemma 1} presented as follows would give the reason.\\

 \indent\textbf{Lemma 1:} \emph{Considering the functions $d^s_{x_i}(\textbf{M},K) $ and $d^d_{x_i}(\textbf{M},K)$ defined in Eq.(\ref{kneighborf}), they are not continuous to the metric matrix $\textbf{M}$. And each non-continuous point changes the $K$ nearest neighbors.}\\
\emph{Proof}:\\
\indent Let us prove the non-continuity of $d^s_{x_i}(\textbf{M},K)$ firstly. We consider two metric matrices $\textbf{M}_1$ and $\textbf{M}_2$. Without loss of generality, we assume that the metric matrix $\textbf{M}_1$ determine the $K$ nearest neighbors of $\textbf{x}_i$ as $\{\textbf{x}^{(1)},\textbf{x}^{(2)},\cdots,\textbf{x}^{(K-1)},\textbf{x}^{(K)}\}$, and the metric $\textbf{M}_2$ determines the $K$ nearest neighbors $\{\textbf{x}^{(1)},\textbf{x}^{(2)},\cdots,\textbf{x}^{(K-1)},\textbf{x}^{(K+1)}\}$. The error between $d^s_{x_i}(\textbf{M}_1,K) $ and $d^s_{x_i}(\textbf{M}_2,K)$ is calculated as follows:
\begin{equation}\label{error}
\begin{split}
&d^s_{x_i}(\textbf{M}_1,K) - d^s_{x_i}(\textbf{M}_2,K)\\
=& \frac{1}{K}\sum_{k=1}^{K-1}Tr(\textbf{X}_{ik}(\textbf{M}_1\!-\!\textbf{M}_2))+\frac{1}{K}Tr((\textbf{X}_{iK}\!-\!\textbf{X}_{i{K+1}})\textbf{M}_1)\\
&\quad\quad\quad+ \frac{1}{K}Tr(\textbf{X}_{iK+1}(\textbf{M}_1-\textbf{M}_2) )
\end{split}
\end{equation}
where $\textbf{X}_{ij} = (\textbf{x}_i-\textbf{x}_j)(\textbf{x}_i-\textbf{x}_j)^T$ and $Tr(\cdot)$ is trace of matrix. If the $d^s_{x_i}(\textbf{M},K)$ is continuous, the error presented in Eq.(\ref{error}) would be zero when $\textbf{M}_1$ approaches $\textbf{M}_2$ $(\textbf{M}_1\rightarrow \textbf{M}_2)$. However, the error equals $Tr((\textbf{X}_{iK}\!-\!\textbf{X}_{i{K+1}})\textbf{M}_1)/K \neq 0$. Therefore, $d^s_{x_i}(\textbf{M},K)$ is not a continuous function with respected to $\textbf{M}$. Let the $\mathcal{S}_i$ in the $d^s_{x_i}(\textbf{M},K)$ be replaced by $\mathcal{D}_i$, the non-continuity of $d^d_{x_i}(\textbf{M},K)$ can be proved in the same way.\\
$\Box$\\
\indent The \textbf{Lemma 1} indicates that when one of the $K$ nearest neighbors is changed, there exists a corresponding non-continuous point $\textbf{M}'$. Therefore, in the searching space, when $d^s_{x_i}(\textbf{M}',K)$ (or $d^d_{x_i}(\textbf{M}',K)$) is continuous at the $\textbf{M}'$, the metric $\textbf{M}'$ would not change the $K$-nearest neighbors. As we know, the gradient decent method can not across the non-continuous point in the searching procedure. So the $K$ nearest neighbors are not changed in the procedure of the gradient decent method, which means there is no selection of $K$ nearest neighbors for the inquiry sample $\textbf{x}_i$. However, the proposed method in Eq.(\ref{imtco2}) requires a selection of the $K$ nearest neighbors from the similarity set (or dissimilarity set). Therefore, the gradient descent method can not achieve the goal of the proposed method.\\
 \subsection{Formulation of the Adaptive Nearest Neighbour (ANN)}
 \indent In this section, we design a continuous function to estimate the $d^s_{x_i}(\textbf{M},K)$ and $d^d_{x_i}(\textbf{M},K)$ defined in Eq.(\ref{kneighborf}). By using the continuous function, we can obtain a continuous model named adaptive nearest neighbour (ANN). Before doing this, we intoruce two useful lemmas as follows.\\

 \textbf{Lemma 2.} \emph{Given a series of numbers $\left\{a_i\right\}_{i=1}^n$, without loss of generality, they are listed in ascending order, i.e., $ a_1\leq a_2\leq\cdots\leq a_n$. Considering the function}
 \begin{equation}\label{lm1}
 \begin{split}
&b\left(\gamma\right) = -\frac{1}{\gamma}\\ln\left( \frac{\sum_{i=1}^ne^{-\gamma a_i}}{n}\right)\\
 \end{split}
 \end{equation}
\emph{, there exist following relationships:}
 \begin{equation}\label{lm11}
 \begin{split}
 \lim_{\gamma\rightarrow 0}b\left(\gamma\right)=\sum_{i=1}^n\frac{a_i}{n},\lim_{\gamma\rightarrow+\infty}b\left(\gamma\right)=a_1,\lim_{\gamma\rightarrow-\infty}b\left(\gamma\right)=a_n
 \end{split}
 \end{equation}

\emph{Proof}:\\
\indent Firstly, we prove the conclusion that $\lim_{\gamma\rightarrow+\infty}b\left(\gamma\right)=a_1 $.\\
\indent Let $t_i = e^{-\gamma(a_i-a_1)} (i >1)$, $b(\gamma)$ can be rewritten as $b(\gamma)=a_1-\frac{1}{\gamma}\ln(\frac{1+\sum_{i=2}^nt_i}{n})$. Since $\gamma > 0$ and $a_i < a_{i+1}$, there is $1>t_2 \geq t_3 \geq \cdots \geq t_n$. By utilizing the inequality relationship $1\geq t_i\geq t_n$, we can obtain following inequality.
\begin{equation}\label{lem3}
a_1=a_1\!-\!\frac{1}{\gamma}\ln(1)\!\!\leq b\left(\gamma\right)\leq\!\!a_1\!-\!\frac{1}{\gamma}\ln(\frac{1+\sum_{i=2}^nt_n}{n})
 \end{equation}
 Since there is $lim_{\gamma \rightarrow +\infty}\frac{1}{\gamma}\ln(\frac{1+\sum_{i=2}^nt}{n}) = 0$ when $t < 1$, so when $\gamma \rightarrow +\infty$, the upper bound of $b\left(\gamma\right)$ approaches $a_1$. Consequently, we obtain the conclusion that $ b\left(\gamma\right) \rightarrow a_1 (\gamma \rightarrow +\infty)$.\\
\indent Then, we prove the conclusion $\lim_{\gamma\rightarrow-\infty}b\left(\gamma\right)=a_n $. Let $c = -\gamma$, we have
 \begin{equation}\label{lmbian1}
 \begin{split}
&\quad\quad\quad \quad b\left(c\right)=\frac{1}{c}\ln\left( \frac{\sum_{i=1}^ne^{-c (-a_i)}}{n}\right)\\
 \end{split}
 \end{equation}
 Therefore,
 \begin{equation}\label{lm2}
 \quad\quad\lim_{\gamma \rightarrow -\infty}b\left(\gamma\right) = \lim_{c \rightarrow +\infty}b\left(c\right) =- {\min\{(-a_i)\}_{i=1}^n}=a_n
 \end{equation}
 \indent At last, we prove that $\lim_{\gamma \rightarrow 0}b(\gamma) = \frac{1}{n}\sum_{i=1}^na_i$. According to the\emph{ ``L'Hospital rule''} \cite{ko2006mathematical}, there is\\
 \begin{equation}\label{e_proof}
 \begin{split}
 \lim_{\gamma \rightarrow 0}b(\gamma) &= \lim_{\gamma \rightarrow 0} \frac{d\left( \frac{\sum_{i=1}^n e^{-\gamma a_i}}{n}\right)}{d\gamma}/1=\lim_{\gamma \rightarrow 0}\frac{1}{n}\sum_{i=1}^n e^{-\gamma a_i}a_i\\
 &=\frac{1}{n}\sum_{i=1}^n a_i
 \end{split}
 \end{equation}
 \indent
$ \Box$\\
 \textbf{Lemma 3.} \emph{Given a set of numbers listed in ascending order, i.e, $a_1<a_2<\cdots<a_n$ and an integer $K \leq n$, there exist $\gamma_1^*>0$ and $\gamma_2^*<0$ to let $b(\gamma)$ defined in Eq.(\ref{lm1}) to hold following equations.}
 \begin{equation}\label{lemma2}
 \begin{split}
b(\gamma_1^*)=\frac{1}{K}\sum_{k=1}^Ka_k, \quad b(\gamma_2^*)=\frac{1}{K}\sum_{k=1}^Ka_{(n-k+1)}
 \end{split}
 \end{equation}
\textbf{ Proof:} Since the proofs of the two equalities have the same procedure, we only provide the proof of the first equality here. \\
\indent Obviously, there is inequality $\frac{1}{n}\sum_{i=1}^na_i>\frac{1}{K}\sum_{k=1}^Ka_k > a_1$. Since \textbf{ Lemma 2} concludes that $\lim_{\gamma \rightarrow+\infty}b(\gamma) = a_1$ and $\lim_{\gamma \rightarrow 0}b(\gamma) = \frac{1}{n}\sum_{i=1}^na_i$, we can easily find a value $\gamma^{(1)} >0$ to hold following inequality:
 \begin{equation}\label{lemma21}
 \quad\quad b(\gamma^{(1)})\leq \frac{1}{K}\sum_{k=1}^Ka_k \leq b(0)
 \end{equation}
Let us define a function $\hat{b}(\gamma) =  \frac{1}{K}\sum_{k=1}^Ka_k - b(\gamma)$. Obviously, $\hat{b}(\gamma^{(1)}) >0$ and $\hat{b}(0) <0$. Since $ \hat{b}(\gamma)$ is a continuous function with respected to $\gamma$, according to `the existence theorem of zero points' \cite{Herings2004General}, there exits a value $\gamma_1^* \in [0,\gamma^{(1)}]$ to make $\hat{b}(\gamma^*_1) =0$. The first equation in Eq.(\ref{lemma2}) is proved.\\
\indent Similarly, we can prove the second equation in Eq.(\ref{lemma2}).\\
$\Box$\\
\indent\textbf{ Lemma 3} implies that the continuous function $b(\gamma)$ can be used to estimate \emph{the average value of $K$ smallest values (or largest values) }of a set of numbers. Let us define $d^s_{x_i}(\textbf{M},\gamma)$ and $d^d_{x_i}(\textbf{M},\gamma)$ as follows:
\begin{equation}\label{intro_func}
\begin{split}
d^s_{x_i}(\textbf{M},\gamma)&= \frac{1}{-\gamma}\ln\left(\frac{1}{\vert\mathcal{S}_i\vert}\sum_{j\in\mathcal{S}_i}{e^{-\gamma d_\textbf{M}(\textbf{x}_j,\textbf{x}_i)}}\right)\\
\end{split}
\end{equation}
\begin{equation}\label{intro_func22}
\begin{split}
\quad\quad d^d_{x_i}(\textbf{M},\gamma)&= \frac{1}{-\gamma}\ln\left(\frac{1}{\vert\mathcal{D}_i\vert}\sum_{j\in\mathcal{D}_i}{e^{-\gamma d_\textbf{M}(\textbf{x}_j,\textbf{x}_i)}}\right)\\
\end{split}
\end{equation}
\indent In above equations, $\vert \mathcal{S}_i\vert$ and $\vert \mathcal{D}_i\vert$ are the numbers of samples in $\mathcal{S}_i$ and $\mathcal{D}_i$, respectively.\\
 \indent According to \textbf{Lemma 3}, for data sets $\mathcal{S}_i$ and $\mathcal{D}_i$, there are two values $\gamma_1 $ and $\gamma_2 $ to make following equations be established.
 \begin{equation}\label{into_pqeq}
 d^s_{x_i}(\textbf{M},\gamma_1) = d^s_{x_i}(\textbf{M},K), \quad d^d_{x_i}(\textbf{M},\gamma_2) = d^d_{x_i}(\textbf{M},K)
 \end{equation}
 where, $ d^s_{x_i}(\textbf{M},K)$ and $d^d_{x_i}(\textbf{M},K)$ are defined in Eq.(\ref{kneighborf}).\\
\indent By substituting Eq.(\ref{intro_func}), Eq (\ref{into_pqeq}) and Eq.(\ref{kneighborf}) into Eq.(\ref{imtrco1}), we have a new constraint to let the $i$-th inquiry sample be classified correctly. The new constraint is presented as follows
\begin{equation}\label{constaint2}
\begin{split}
&\frac{1}{-\gamma_1}\ln\left(\frac{1}{\vert\mathcal{S}_i\vert}\sum_{j\in\mathcal{S}_i}{e^{-\gamma_1d_\textbf{M}(\textbf{x}_i,\textbf{x}_j)}}\right)< \\ &\quad\quad\quad\quad\quad\quad\quad\frac{1}{-\gamma_2}\ln\left(\frac{1}{\vert\mathcal{D}_i\vert}\sum_{l\in\mathcal{D}_i}{e^{-\gamma_2 d_\textbf{M}(\textbf{x}_i,\textbf{x}_l)}}\right)
\end{split}
\end{equation}
\indent Above constraint can be further transformed. Since the distance function $d_\textbf{M}(\textbf{x}_i,\textbf{x}_l)$ is linear to the metric matrix $\textbf{{M}}$, i.e., $\gamma_2d_\textbf{M}(\textbf{x}_i,\textbf{x}_l) =d_{ (\gamma_2\textbf{M})}(\textbf{x}_i,\textbf{x}_l) $. Let us set $\gamma_2 > 0$, the Eq.(\ref{constaint2}) can be equivalently changed into Eq.(\ref{constaint3}) by adopting $\hat{\textbf{M}}=\gamma_2\textbf{M}$:\\
\begin{equation}\label{constaint3}
\begin{split}
&-\frac{1}{\alpha} \ln\left(\frac{1}{\vert\mathcal{S}_i\vert}\sum_{j\in\mathcal{S}_i}{e^{-\alpha d_{\hat{\textbf{M}}}(\textbf{x}_i,\textbf{x}_j)}}\right)\frac{1}{\gamma_2} < \\ &\quad\quad\quad\quad\quad\quad\quad\quad - \ln\left(\frac{1}{\vert\mathcal{D}_i\vert}\sum_{l\in\mathcal{D}_i}{e^{-d_{\hat{\textbf{M}}}(\textbf{x}_i,\textbf{x}_l)}}\right)\frac{1}{\gamma_2}
\end{split}
\end{equation}
where $\alpha = \frac{\gamma_1}{\gamma_2}$, and $\gamma_2 > 0$. Obviously, eliminating $\gamma_2$ does not impact the inequality. Since $\gamma_2$ determines the number of the nearest neighbors in $\mathcal{D}_i$, the number of the nearest neighbors of $\textbf{x}_i$ in $\mathcal{D}_i$ can be determined adaptively. However, considering the non-linear of the loss function, $\gamma_2$ can not be eliminated here. Thus, the $\gamma_2$ only plays the rule to adjust the sensitivity of the loss function, and $\alpha$ plays the rule to make the number of nearest neighbors in $\mathcal{D}_i$ equal to that in $\mathcal{S}_i$.\\
\indent By adopting the constraint Eq.(\ref{constaint3}), the non-continuous empirical risk minimization problem in Eq.(\ref{imtco2}) is transformed into the continuous one presented as follows:
  \begin{equation}\label{optim_continous}
 \begin{split}
  \min_{\textbf{M}\succeq 0} \sum_{i=1}^N & \ell\left(-\frac{1}{\alpha} \ln\left(\frac{1}{\vert\mathcal{S}_i\vert}\sum_{j\in\mathcal{S}_i}{e^{-\alpha d_{{\textbf{M}}}(\textbf{x}_i,\textbf{x}_j)}}\right)\frac{1}{\gamma}+\right.\\
  &\quad\quad\quad\quad\quad \left. \ln\left(\frac{1}{\vert\mathcal{D}_i\vert}\sum_{l\in\mathcal{D}_i}{e^{-d_{{\textbf{M}}}(\textbf{x}_i,\textbf{x}_l)}}\right)\frac{1}{\gamma}\right) \\
  &\quad\quad\quad\quad\quad\quad\quad\quad\quad+ \lambda \sum_{i=1}^N\sum_{j\in \mathcal{S}_i}d_\textbf{M}(\textbf{x}_i,\textbf{x}_j)
 \end{split}
 \end{equation}
  where $\gamma >0$ plays the rule to adjust the loss function and $\alpha$ is used to balance the numbers of nearest neighbors of $\textbf{x}_i$ in $\mathcal{S}_i$ and $\mathcal{D}_i$, $\lambda$ is the coefficient of the regularization term.\\
\subsection{Optimization}
 In this section, we solve the proposed method by the gradient descent method. Let us denote the objective function of the proposed method in Eq.(\ref{optim_continous}) as $J(\textbf{M})$. With the help of Eq.(\ref{intro_func}), $J(\textbf{M})$ can be rewritten as
\begin{equation}\label{jj_obj}
\begin{split}
J(\textbf{M}) \!=\!\!&\sum_{i=1}^N\ell(\frac{1}{\gamma} (d^d_{x_i}(\textbf{M},1)\!-\! d^s_{x_i}(\textbf{M},\alpha))) \!+\! \lambda \!\!\sum_{i=1}^N\!\!\sum_{j\in \mathcal{S}_i}d_\textbf{M}(\textbf{x}_i,\textbf{x}_j)
\end{split}
\end{equation}
The gradient of $J(\textbf{M})$ associated with $\textbf{M}$ is calculated as:
\begin{equation}\label{dml_ipr}
\begin{split}
\frac{\partial J(\textbf{M})}{\partial \textbf{M}}=&\sum_{i=1}^N\left(\xi_i(\frac{\partial d^s_{x_i}({\textbf{M}},\alpha)}{\partial \textbf{M}}-\!\!\frac{\partial d^d_{x_i}({\textbf{M}},1)}{\partial \textbf{M}}) + \lambda\sum_{j\in \mathcal{S}_i} \textbf{X}_{ij}\right)
\end{split}
\end{equation}
where,
\begin{equation}\label{mdl_xi}
\xi_i = \ell'(\frac{1}{\gamma}d^s_{x_i}({\textbf{M}},\alpha)-\frac{1}{\gamma} d^d_{x_i}({\textbf{M}},1))
\end{equation}
and,
\begin{equation}\label{mdl_itccc}
\begin{split}
\frac{\partial d^s_{x_i}({\textbf{M}},\alpha)}{\partial \textbf{M}} = \sum_{j \in \mathcal{S}_i} r^s_{ij}\textbf{X}_{ij},\quad \frac{\partial d^d_{x_i}({\textbf{M}},1)}{\partial \textbf{M}} = \sum_{l \in \mathcal{D}_i} r^d_{il}\textbf{X}_{il}\\
\end{split}
\end{equation}
 where,
\begin{equation}\label{coeiff}
\begin{split}
r^s_{ij} = \frac{e^{-\alpha d_{\textbf{M}}(\textbf{x}_i,\textbf{x}_j)}}{\sum_{j\in \mathcal{S}_i}e^{-\alpha d_{\textbf{M}}(\textbf{x}_i,\textbf{x}_j)}}, \quad r^d_{il} =\frac{e^{- d_{\textbf{M}}(\textbf{x}_i,\textbf{x}_l)}}{\sum_{l\in \mathcal{D}_i}e^{-d_{\textbf{M}}(\textbf{x}_i,\textbf{x}_l)}}
\end{split}
\end{equation}
and $\textbf{X}_{ij} = (\textbf{x}_i-\textbf{x}_j)(\textbf{x}_i-\textbf{x}_j)^T$, $\textbf{X}_{il} = (\textbf{x}_i-\textbf{x}_l)(\textbf{x}_i-\textbf{x}_l)^T$.
\indent Look at the Eq.(\ref{coeiff}), $r^s_{ij}$ and $r^d_{il}$ are the coefficients to $\textbf{X}_{ij}$ and $\textbf{X}_{il}$, respectively. It is easy to find that the $r^s_{ij}$ and $r^d_{il}$ are two softmax functions related to each inquiry $\textbf{x}_i$. As we know the softmax function is a smooth proxy of the $ \arg\max$ function which is often adopted as the classification boundary. Here, softmax function can be interpreted as finding the soft $K$ nearest neighbors of $\textbf{x}_i$ from $\mathcal{S}_i$ and $\mathcal{D}_i$. This procedure is similar to the direct selection of $K$ nearest neighbors by sorting distance $\{d_{\textbf{M}}(\textbf{x}_i,\textbf{x}_j)\}_{j\in \mathcal{S}_i}$, which makes $d^s_{x_i}({\textbf{M}},K)$ and $d^d_{x_i}({\textbf{M}},K)$ be non-continuous and the problem be hard to solve. However, by using the coefficients in Eq.(\ref{coeiff}), those drawbacks are avoided.\\
\indent After giving the computation of the gradient of the objective function, the proposed method can be solved iteratively. Suppose the $\eta_s$ and $\textbf{M}^s$ are the step-size and the projection matrix at the $s$-th iteration, respectively. Then, the projection matrix at $s+1$ iteration is updated as
\begin{equation}\label{grad_up}
\textbf{M}^{s+1} =\psi(\textbf{M}^s - \eta_s\frac{\partial J_1(\textbf{M})}{\textbf{M}}|_{\textbf{M} = \textbf{M}^s})
\end{equation}
where $\psi(\cdot)$ is defined in Eq.(\ref{sdp_proj}) to make matrix $\cdot$ be negative-positive definite.\\
\begin{equation}\label{sdp_proj}
\psi(\textbf{M}) = \sum_{i=1}^r\sigma_i \textbf{u}_i\textbf{u}^T_i
\end{equation}
where $\{(\sigma_i,\textbf{u}_i)\}_{i=1}^r$ are the $r$ positive eigen-values and corresponding eigen-vectors of $\textbf{M}$.\\
\indent The details of the solution are presented in Algorithm~{\ref{algexample}}.\\
\section{The Relationship Between ANN and the Existing Metric Learning Algorithms}
\label{relationship_ann_other}
\indent In this section, we would discuss the relationships between the proposed ANN and the existing methods including the large margin nearest neighbors (LMNN), neighbourhood components analysis (NCA) and the naive form of pairwise-based distance metric learning methods \cite{hu2017sharable,ye2019learning}.\\
\subsection{The Connection between ANN and LMNN}
\indent Firstly, we introduce the Lemma 4 which claims that the ANN presented in Eq.(\ref{optim_continous}) is a convex model when $\alpha < 0$.\\
\indent \textbf{Lemma 4}: When $\alpha < 0$, the proposed ANN described in Eq.(\ref{optim_continous}) is a convex optimization problem.\\
\emph{\textbf{Proof}}: Firstly, we introduce a function $S(\textbf{t}) = \ln(\frac{1}{d}\sum_{i=1}^d e^{t_i})$ where $\textbf{t}=[t_1,t_2,\cdots,t_d]\in \mathbb{R}^d$. Then, we prove the function $S(\textbf{t})$ is convex. Suppose $0<\tau <1$ and two vectors $\textbf{t}^1 =[t^{(1)}_1,t^{(1)}_2,\cdots,t_d^{(1)}]$ and $\textbf{t}^2=[t_1^{(2)},t_2^{(2)},\cdots,t_d^{(2)}]$, we have
\begin{equation}\label{lemma4_eq1}
\begin{split}
&\quad\quad S(\tau\textbf{t}_1+(1-\tau)\textbf{t}_2) = \ln(\sum_{i=1}^d e^{\tau t^{(1)}_i+(1-\tau)t^{(2)}_i})-\ln(d)\\
&\leq  \ln((\sum_{i=1}^d e^{\tau t^{(1)}_i})(\sum_{i=1}^d e^{(1-\tau)t^{(2)}_i}))-\ln(d)\\
\end{split}
\end{equation}
\begin{equation}\label{lemma4_eq2}
\begin{split}
&= \ln(\sum_{i=1}^d e^{\tau t^{(1)}_i})+\ln(\sum_{i=1}^d e^{(1-\tau)t^{(2)}_i})\!-\!(1\!-\!\tau)\ln(d)\!-\!\tau \ln(d)\\
&\leq  \ln((\sum_{i=1}^d e^{t^{(1)}_i})^{\tau})+\ln((\sum_{i=1}^d e^{t^{(2)}_i})^{(1\!-\!\tau)}) \!-\!(1\!-\!\tau)\ln(d) \!-\!\tau \ln(d)\\
&=\tau S(\textbf{t}_1)+(1-\tau)S(\textbf{t}_2)
\end{split}
\end{equation}
The inequality in Eq.(\ref{lemma4_eq1}) is hold by the $\sum{a_ib_i} < \sum{a_i}\sum{b_i}$ when $a_i >0$ and $b_i>0$, and the  inequality in Eq.(\ref{lemma4_eq2}) is hold by $\sum_i {a_i^{\tau}} < (\sum_i{a_i})^{\tau}$ when $\tau <1$.\\
\indent Let $\textbf{t}^{(1)}_i= [\alpha d_\textbf{M}(\textbf{x}_i,\textbf{x}^1_j),\cdots,\alpha d_\textbf{M}(\textbf{x}_i,\textbf{x}^{\vert \mathcal{S}_i\vert}_j)] \in \mathbb{R}^{\vert \mathcal{S}_i\vert}$, where $\{\textbf{x}^1_j,\cdots,\textbf{x}_j^{\vert \mathcal{S}_i\vert}\} = \mathcal{S}_i$, therefore, $S(\textbf{t}^{(1)}_i)$ is convex with respective to $\textbf{M}$. Since $\alpha < 0$, $d^s_{{x}_i}(\textbf{M},\alpha) = -\frac{1}{\alpha} S(\textbf{t}^{(1)}_i)$ is convex.\\
\indent Then, let $\textbf{t}^{(2)}_i= [d_\textbf{M}(\textbf{x}_i,\textbf{x}^1_l),\cdots, d_\textbf{M}(\textbf{x}_i,\textbf{x}^{\vert \mathcal{D}_i\vert}_l)] \in \mathbb{R}^{\vert \mathcal{D}_i\vert}$, where $\{\textbf{x}^1_l, \cdots,\textbf{x}^{\vert \mathcal{D}_i\vert}_l\} = \mathcal{D}_i$, $S(\textbf{t}^{(2)}_i)$ is convex with respective to $\textbf{M}$. So, $-d^d_{{x}_i}(\textbf{M},1) = S(\textbf{t}^{(2)}_i)$ is convex.\\
\indent Combining the conclusions above, the function $d^s_{{x}_i}(\textbf{M},\alpha)-d^d_{{x}_i}(\textbf{M},1)$ is convex. Due to the loss function $\ell(\cdot)$ is also a convex function, the objective function
in Eq.(\ref{optim_continous}) is convex. Since the searching space $\{\textbf{M}|\textbf{M}\succeq 0\}$ is a convex set, therefore the model in Eq.(\ref{optim_continous}) is a convex optimization problem.\\
$\Box$
\\
 \indent Lemma 4 proves the ANN is convex when $\alpha<0$ above. Now, we explain that, when $\alpha \rightarrow -\infty$ and similarity set $\mathcal{S}_i = \mathcal{P}_i$, the model described in Eq.(\ref{optim_continous}) is equivalent to the model of LMNN.\\
\indent When $\alpha \rightarrow -\infty$, $\gamma_1$ approaches $-\infty$ (due to $\gamma_1= \alpha\gamma_2$). Thus, the function $d^s_{{x}_i}(\textbf{M},\gamma_1)= \sup\{d_\textbf{M}(\textbf{x}_i,\textbf{x}_j)| \textbf{x}_j \in \mathcal{S}_i\} = d^{max}_i$. In this case, the objective of ANN is to let most of the samples in $\mathcal{D}_i$ out of the neighborhood $\mathcal{A}_i = \{\textbf{x}| d_\textbf{M}(\textbf{x}_i,\textbf{x})< d_i^{max}\}$. This is consistent with the goal of the LMNN. Since the \textbf{Lemma 4} proves the model described in Eq.(\ref{optim_continous}) is convex when $\alpha <0$. Therefore, we can claim that the model in Eq.(\ref{optim_continous}) is equivalent to the LMNN. \\
\indent Although ANN($\alpha=-\infty$) achieves the same goal of LMNN, the searching space is much boarder than that of LMNN. In literature \cite{}, the LMNN equals the SVM whose decision boundary is determined by the support vectors. However, the support vectors are the samples on the enveloping line of the regions determined by the samples of each class. Therefore, not only the nearest samples from the decision boundary but also the farthest samples from the decision boundary are considered. In the ANN, only the nearest samples from the decision boundary are considered as the constraints of the model. Therefore, the ANN ($\alpha=-\infty$) has much boarder searching space than that of LMNN.\\
\indent When $\alpha < 0 \& \alpha \neq -\infty$, the ANN is more robust to noise than LMNN. That is because the constraint $\frac{1}{K}\sum_{k=1}^Kd^{[\vert\mathcal{S}_i\vert -k+1]}_\textbf{M}(\textbf{x}_i,\textbf{x}_j)< \frac{1}{K}\sum_{k=1}^Kd^{[k]}_\textbf{M}(\textbf{x}_i,\textbf{x}_j)$ involves the average distance which can reduce the effect of noises which may produce bad imposters.\\
\indent \textbf{Remark 4}: The computational complexity of ANN is much less than that of LMNN. In the solution procedure of the ANN, the number of computations of $d_\textbf{M}(\textbf{x}_i,\textbf{x}_j)$ is about $N(N-1)$ in each iteration, while in LMNN there are $K$ target neighbors and about $\vert \mathcal{P}_i\vert N^2(C-1)/C$ times distance computation involved, where $\vert \mathcal{P}_i\vert$ is the number if target neighbors for each inquiry sample. Commonly, there are $\vert \mathcal{P}_i\vert\geq 3$ and $C\geq 2$. Let us roughly suppose that both ANN and LMNN have equal number of iterations, we could obtain that the running speed of ANN is about ${\vert \mathcal{P}_i\vert}$ than that of LMNN.\\
\subsection{The Connection between ANN and NCA}
\label{pnca}
\indent \textbf{Remark 5}: When we set the parameter as $\alpha = 1$ and the loss function as $\ell(x)=x$, the ANN described in Eq.(\ref{optim_continous}) is equivalent to the neighbourhood components analysis (NCA).\\
\indent For each inquiry sample $\textbf{x}_i$, the objective of NCA is to maximize $p_i$ defined in Eq.(\ref{NCA_pp1}).
Let us set the loss function in ANN as $\ell(x) = x$, then we can equivalently transform the loss of $\textbf{x}_i$ of ANN in following procedure
\begin{equation}\label{Eq_remark5}
\begin{split}
  &\min_{\textbf{M}\succeq 0} \ell(-\frac{1}{\alpha}\ln(\frac{1}{\vert S_i\vert}\sum_{j \in \mathcal{S}_i} e^{-\alpha d_\textbf{M}(\textbf{x}_i,\textbf{x}_j)}) +\\
  & \quad\quad\quad\quad\quad\quad\quad\frac{1}{\vert D_i\vert}\ln(\sum_{j \in {\mathcal{D}}_{i}} e^{-d_\textbf{M}(\textbf{x}_i,\textbf{x}_j)}))\\
\Leftrightarrow& \min_{\textbf{M}\succeq 0}\frac{\sum_{j \in {\mathcal{D}}_{i}} e^{-d_\textbf{M}(\textbf{x}_i,\textbf{x}_j)}}{(\sum_{j \in \mathcal{S}_{i}} e^{-\alpha d_\textbf{M}(\textbf{x}_i,\textbf{x}_j)})^{1/\alpha}}\\
\end{split}
\end{equation}
The Eq.(\ref{Eq_remark5}) is equivalent to the optimization problem presented as followings.
\begin{equation}\label{weight_NCA}
  \max_{\textbf{M}}\frac{(\sum_{j \in \mathcal{S}_{i}} e^{-\alpha d_\textbf{M}(\textbf{x}_i,\textbf{x}_j)})^{1/\alpha}}{{\sum_{j \in {\mathcal{D}}_{i}} e^{-d_\textbf{M}(\textbf{x}_i,\textbf{x}_j)}}+{(\sum_{j \in \mathcal{S}_{i}} e^{-\alpha d_\textbf{M}(\textbf{x}_i,\textbf{x}_j)})^{1/\alpha}}}
\end{equation}
\indent At last, by considering all the samples in the training data set, we obtain a new model called as parameterized neighbourhood components analysis (PNCA) depicted as follows: \\
\begin{equation}\label{para_nac}
\max_{\textbf{M}\succeq 0} \sum_{i=1}^N\frac{(\sum_{j \in \mathcal{S}_{i}} e^{-\alpha d_\textbf{M}(\textbf{x}_i,\textbf{x}_j)})^{1/\alpha}}{{\sum_{j \in {\mathcal{D}}_{i}} e^{-d_\textbf{M}(\textbf{x}_i,\textbf{x}_j)}}+{(\sum_{j \in \mathcal{S}_{i}} e^{-\alpha d_\textbf{M}(\textbf{x}_i,\textbf{x}_j)})^{1/\alpha}}}
\end{equation}
\indent Let us select the $\mathcal{S}_i$ as the set of all the samples in the class ${y_i}$ except $\textbf{x}_i$, and $\mathcal{D}_i$ as the set of all the samples in classes different from $y_i$. When we set $\alpha = 1$, the objective function in the above formulation presented in Eq.(\ref{para_nac}) becomes the formulation of NCA. Therefore, the NCA is a special case of ANN. \\
\indent Actually, the objective in Eq.(\ref{weight_NCA}) can be still seen as the probability of $\textbf{x}_i$ being classified correctly. And the parameter $\alpha$ plays the rule to balance the numbers of samples in the set $\mathcal{S}_{i}$ and $\mathcal{D}_i$. By adjusting the parameter $\alpha$, the problem of ignoring some dissimilar samples in NCA would overcome. As the result, more discriminant information are considered.\\
\subsection{The Connection between ANN and the Pairwise Constraint Metric Learning}
\indent In this subsection, we present the connection between the ANN and the pairwise constraint metric learning (PML) algorithms \cite{davis2007information,mignon2012pcca,hu2017sharable}. Since there are plenty of variants of PML algorithms, we only consider the general form of the PML. Without loss of generality, we also ignore the loss functions used in both ANN and PML.\\
\indent The general formulation of a pairwise constraint metric learning algorithm is presented as follows:
\begin{equation}\label{Pw_B_ML}
\begin{split}
&\quad\quad \text{minimize} \quad\!\!\!\Omega(\textbf{M}) \\
s.t.\quad& d_\textbf{M}(\textbf{x}_i,\textbf{x}_j) < u, (\textbf{x}_i,\textbf{x}_j) \in \mathcal{S};\\
& d_\textbf{M}(\textbf{x}_i,\textbf{x}_j) > l, (\textbf{x}_i,\textbf{x}_j) \in \mathcal{D};\\
& {\textbf{M}\succeq 0}.
\end{split}
\end{equation}
where $\Omega(\textbf{M})$ is the regularization term of $\textbf{M}$, $\mathcal{S}=\{(\textbf{x}_i,\textbf{x}_j)|y_i = y_j, i \neq j\}$, $\mathcal{D} = \{(\textbf{x}_i,\textbf{x}_j), y_i \neq y_j\}$ and $0 < u \leq l$. Obviously, the searching space determined by the pairwise constraints is a subset of that of following constraints
 \begin{equation}\label{pair_cons}
   \mathcal{P}\!=\!\bigcap_{i=1}^N \{\textbf{M}| d_\textbf{M}(\textbf{x}_i,\textbf{x}_j )< u\!-\!l\!+\!d_\textbf{M}(\textbf{x}_i,\textbf{x}_l), \textbf{x}_j \in \mathcal{S}_i ,\textbf{x}_l \in \mathcal{D}_i \}
\end{equation}
where $\mathcal{S}_i$ and $\mathcal{D}_i$ are symbols used in ANN.
Since the pairwise constraint metric learning algorithms capture global information, we consider the naive form of ANN with $\alpha = -\infty$ and $\gamma = +\infty$. Its formulation is presented as follows:
\begin{equation}\label{ANN_ML}
\begin{split}
&\quad \text{minimize}\quad\!\!\!\Omega(\textbf{M}) \\
s.t.\quad&d^{[\vert \mathcal{S}_i\vert]}_\textbf{M}(\textbf{x}_i,\mathcal{S}_i)< d^{[1]}_\textbf{M}(\textbf{x}_i,\mathcal{D}_i);\\
& {\textbf{M}\succeq 0}, \quad\quad i = 1,2,\cdots,N.
\end{split}
\end{equation}
 The searching space of ANN in Eq.(\ref{ANN_ML}) is presented as follows:
 \begin{equation}\label{pair_cons1}
   \mathcal{A} =\bigcap_{i=1}^N\{\textbf{M}| d^{[\vert S_i\vert]}_\textbf{M}(\textbf{x}_i,\mathcal{S}_i)< d^{[1]}_\textbf{M}(\textbf{x}_i,\mathcal{D}_i)\}
\end{equation}
Obviously, when $ u-l = 0$, there is $\mathcal{P} = \mathcal{A}$. Since the searching space determined by Eq.(\ref{Pw_B_ML}) is a subset of $\mathcal{P}$, so the searching space of pairwise constraint metric learning is the subset of that of ANN.\\
\indent \textbf{Remark 6:} The naive form of the pairwise constraint metric learning algorithm is a special case of that of ANN. The only difference is that the searching space of the pairwise constraint metric learning algorithms is a small subset of that of ANN. Compared with ANN, the pairwise metric learning methods are not suitable to deal with the classification task.\\
%
%
%
\begin{table}[!htbp]
\renewcommand{\arraystretch}{1.2} 
\setlength{\tabcolsep}{12pt}  
\centering
\caption{The Details of Datasets.}
\label{data_set}
\footnotesize
\begin{tabular}{|c||c|c|c|}
\toprule
Data set & \# Classes& \#Examples& \#Features\\
\midrule
Australian&2&690&14\\
Cars&2&392&8\\
Ecoli&8&336&343\\
German &2&1,000&20\\
Glass&6&214&9\\
Iris&3&150&4\\
Isolet&2&1,560&617\\
Monk1&2&432&6\\
Solar&6&323&12\\
Vehicle&4&846&18\\
Wine&3&178&13\\
Pendigits&10&10,992&16\\
Coil20&20&1,440&1024\\
Letter&26&20,000&16\\
Usps&10&9,298&256\\
\bottomrule
\end{tabular}
\end{table}
\section{Numerical Experiments}
\indent In this section, we evaluate the proposed metric learning method in four aspects. The first aspect is to explore how the parameters $\alpha$ and $\gamma$ affect the performance of the proposed method. The second aspect is about the comparison of classification accuracy with state-of-the-art methods. At last, the running time of different methods is compared.
\subsection{Data Set Description and Experimental Settings}
We evaluate the proposed ANN on the 15 data sets which are widely adopted to evaluate the performance of machine learning algorithms. All of those data sets come from the UCI Machine learning Repository\footnote{Available at http://archive.ics.uci.edu/ml/datasets.html} and LibSVM\footnote{https://www.csie.ntu.edu.tw/~cjlin/libsvm/}. Since the feature values in some data sets are very large, we normalize them by subtracting the mean and dividing the standard deviation for each feature. The scales of those data sets range from 178 to 20000, their dimensions are various from 4 to 1024, and the number of classes is various from $2$ to $26$. The details of the data sets are presented in Table {\ref{data_set}}. For the data sets whose feature numbers are larger than 150, we utilize principal components analysis (PCA) to reduce the number of their dimensions to $150$.\\
\indent As discussed in section {\ref{relationship_ann_other}}, by adjusting the positive and negative of $\alpha$, the proposed method ANN can be changed into different methods, We use symbol $ANN^{+}$ and $ANN^{-}$ to represent the cases of ANN with $\alpha >0$ and $\alpha<0$, respectively. Since the parameterized neighbourhood component analysis (PNCA) presented in Eq.(\ref{para_nac}) is also derived from our proposed method, we also evaluate its performance in this section.
\subsection{Parameter Determination}
\begin{figure*}[t]
 \centering
\includegraphics[width=0.8\linewidth]{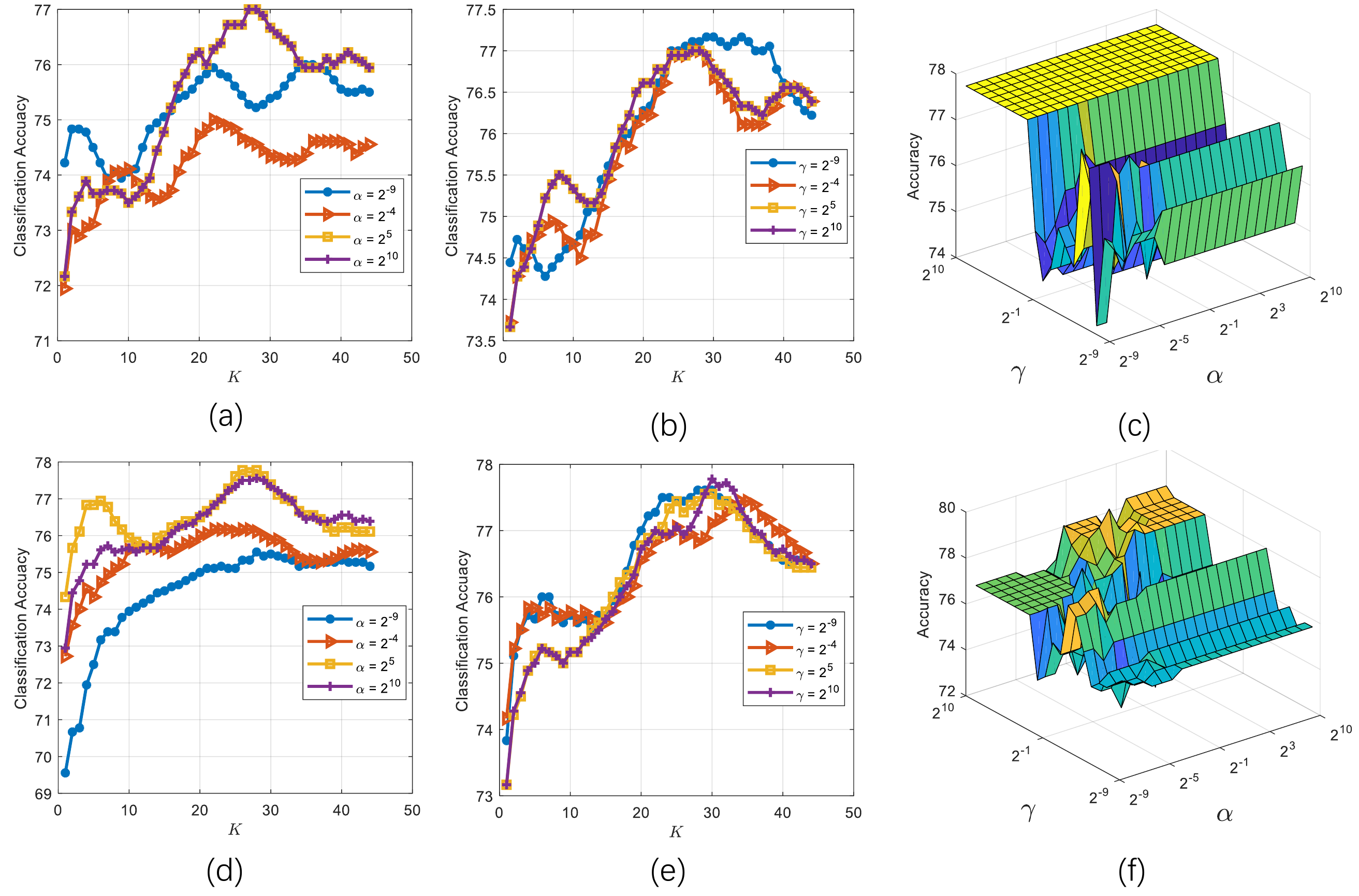}
\caption{The classification results of different parameters ($\alpha$,$\gamma$,$K$) of $ANN^{+}$ and $ANN^{-}$ on the `German' data set. The first row is about $ANN^{-}$, and the second row is about $ANN^{+}$. (a)(d) is $re(\alpha,K)$, (b)(d) is $re(\gamma,K)$, and (c)(e) is $re(\alpha,\gamma)$.}
\label{figkk_1}
\end{figure*}

 In this section, we explore how the parameters affect the methods $ANN^+$ and $ANN^-$. There are two parameters $\alpha$ and $\gamma$  in the two methods. Since we claim that $\gamma$ is related to the number of the nearest neighbors in the dissimilar set $\mathcal{D}_i$, so we also explore the behavior of different $K$ in the $K$-NN used to output the classification results under the learned metric. Thus, in terms of the classification result of the proposed method, three parameters need to consider here, i.e, $\alpha$, $\gamma$, and $K$. We use the symbol $re(\gamma,\alpha,K)$ to denote the classification result to the three parameters. For better illustration, we present the results in three ways as follows.
\begin{itemize}
  \item Firstly, we report the accuracy $re(\gamma,K)$ which represents the best classification result at $\gamma$ and $K$ by adjusting parameter $\alpha$. This would show the relationship between parameters $\gamma$ and $K$;
  \item Secondly, we report the accuracy $re(\alpha,K)$ which represents the best classification result at $\alpha$ and $K$ by adjusting parameter $\gamma$. This would show the relationship between parameters $\alpha$ and $K$;
  \item Thirdly, we report the accuracy $re(\alpha,\gamma)$ which represents the best classification result at $\alpha$ and $\gamma$ by adjusting parameter $K$. this would show the relationship between parameters $\gamma$ and $\alpha$ ;
\end{itemize}
\indent In the experiments, the values of $\alpha$ and $\gamma$ are set as $\{2^{-9},2^{-8},\cdots,2^{10}\}$ (or $\{-2^{-9},-2^{-8},\cdots,-2^{10}\}$), and $K$ is set as $\{1,2,\cdots,50\}$. Since the curves of $re(\gamma,K)$ and $re(\alpha,K)$ are non-smooth with respect to $K$, so we report the smooth results as
\begin{equation}\label{results}
\begin{split}
 \hat{re}(\gamma,K) &= \sum_{i=1}^5 re(\gamma,K+i)/5\\
\hat{re}(\alpha,K) &= \sum_{i=1}^5 re(\alpha,K+i)/5\\
\end{split}
\end{equation}
Due to the limit of space, we only provide the results of the data set `\emph{German}'. For $ANN^{-}$, the similarity set of each inquiry sample is constructed by selecting the $10$ nearest samples under Euclidean metric. For $ANN^{+}$, the similarity set of each inquiry sample is constructed by all the samples in the same class of the inquiry sample. The results are shown in Fig.{\ref{figkk_1}}.\\
\indent As seen from Fig.{\ref{figkk_1}}, we can conclude following conclusions.
\begin{itemize}
  \item As seen from the Fig. \ref{figkk_1} (a) and (d), the curves corresponding to different values of $\alpha$ in each figure have the same variation tendency, but there is a slight difference between the figure (a) and the figure (b). In the figure (a), the curve summit of $\alpha = 2^{5}$ ($\alpha = 2^{10}$) is delayed compared with the curves of $\alpha = 2^{-4}$ and $\alpha = 2^{-9}$, while there is no similar observation found in figure (b). That may be because that the figure (a) is about $ANN^{-}$ which considers the farthest neighbors in similarity set $\mathcal{S}_i$, while the figure (b) is about $ANN^{+}$ which considers the nearest neighbors in the similar set $\mathcal{S}_i$. Thus, for $ANN^{-}$, larger $\alpha$ means larger $K$ used in $K$-NN, while for $ANN^{+}$, larger $\alpha$ means smaller $K$ used in the $K$-NN algorithm. That is why the curves of $ANN^{-}$ have a delay when $\alpha$ is large.
  \item As seen from the Fig \ref{figkk_1} (b) and (e), for both $ANN^+$ and $ANN^-$, the curves corresponding to different values of $\gamma$ have the same variation tendency, and the summits of curves corresponding to different $\gamma$ are attended near the same value of $K$. Since we claim that $\gamma$ is closely related to the number of nearest neighbors in $K$-NN rule, this trend indicates that the proposed method can adaptively determine the $K$ independent to the value of $\gamma$. This observation is consistent with the Eq.(\ref{constaint3}) which is the continuous decision boundary of $K$-NN classifier. Let us replace the parameter $\gamma_2$ in Eq.(\ref{constaint3}) by $\gamma$.
 Obviously, in the inequality, both sides have the term $\frac{1}{\gamma}$, thus the parameter $\gamma$ does not change the inequality. However, $\gamma$ makes a difference in the loss function since the loss function is non-linear. Therefore, by adjusting the value $\gamma$, the loss function is changed to obtain the most appropriate $\textbf{M}$ which would lead to the best classification result. But in the training procedure, the number of the nearest neighbors considered is not changed.
  \item As seen from the Fig. \ref{figkk_1} (c) and (f), we can find that the correlation between $\gamma$ and $\alpha$ in the region contains the best result is very small. Actually, this phenomenon can be also found in other data sets. Since this observation can be found in other data sets, which means the parameters $\gamma$ and $\alpha$ can be tuned separately. This implies that the burden of tuning $\gamma$ and $\alpha$ are not too heavy.
\end{itemize}
\subsection{Classification Accuracy}
\indent In this section, we evaluate the proposed methods on 15 data sets. For those data sets, each of them are split into 70/30 partition for training and testing for 30 times, and the average classification results are reported.\\
\indent We adopt a series of state-of-the-art methods as comparison. They are large margin nearest neighbor (LMNN), information theoretic metric learning (ITML) \cite{davis2007information}, local distance metric learning (LDML) \cite{guillaumin2009you}, sparse component metric learning (SCML) \cite{scml14}, BoostMetric \cite{Shen2009Positive}, neighbourhood components analysis (NCA) \cite{Goldberger2004Neighbourhood}, geometric mean metric learning (GMM) \cite{zadeh2016geometric} and regressive virtual metric learning (RVML) \cite{perrot2015regressive}, etc. \\
\indent Our methods include $ANN^{+}$, $ANN^{-}$ and PNCA. In $ANN^{-}$, for each point $\textbf{x}_i$, the similarity set $\mathcal{S}_i$ is constructed by selecting the $10$ nearest neighbors of $\textbf{x}_i$ from the class $y_i$ under the Euclidean metric, and the dissimilarity set $\mathcal{D}_i$ is constructed by all of the samples from the different classes from $y_i$. In $ANN^{+}$ and PNCA, for each inquiry point $\textbf{x}_i$, the similarity set is constructed by all of the samples in the class $y_i$ except $\textbf{x}_i$. And the dissimilarity set is constructed by all of the samples in the classes different from $y_i$. The parameter $\lambda$ in $ANN^{+}$ and $ANN^{-}$ is set as $1/(N^2)$. The $\alpha$ in $ANN^{+}$ and PNCA is tuned in the grid of $\{2^{-8},2^{-7},\cdots,2^{10}\}$, and the $\alpha$ in $ANN^{-}$ is tuned at the searching grid of $\{-2^{-8},-2^{-7},\cdots,-2^{10}\}$. Since $ANN^{+}$ and PNCA are non-convex optimization problems, we set the initial searching point as $\textbf{M}_0 = \frac{\textbf{I}}{\sqrt{N}}$, where $\textbf{I}$ is the identity matrix. \\
\indent In LMNN, $\lambda$ is tuned at the searching grid $\{0.1,0.2,\cdots,0.9\}$, the target neighbors' number is searched in grid $\{4,\cdots,10\}$. For GMML, the parameter $t$ is tuned in grid $\{0.1,0.2,\cdots,0.9\}$. For ITML, the parameter $\gamma$ is tuned in the grid $\{0.25,0.5,0.7,0.9\}$. All of the other parameters are set as default. Those tuned parameters are determined with 5-fold cross validation. After the metric learning step, the result of $ANN^+$, $ANN^{-1}$, PNCA, and NCA are output by the $K$-nearest neighbor classifier with best $K$ in $\{1,4,7,\cdots,46\}$. For other methods, we report the best results output by $K$-NN with $K\in \{1,2,\cdots,10\}$. The result are shown in Table \ref{ta1_result}.\\
\indent As seen from Table \ref{ta1_result}, we can obtain following conclusions.
\begin{itemize}
  \item In most of the methods, the proposed methods, including $ANN^{+}$, $ANN^{-}$ and PNCA have achieved the best result. This indicates the superiority of the proposed methods.
  \item In some data sets, $ANN^{+}$ has obtained the best results, and in other data sets, $ANN^{-}$ has obtained the best results. Considering that the $ANN^{+}$ is a non-convex model, whose results may be not optimal. So this is consistent to the assumption that $ANN^{-}$ has a boarder searching space than $ANN^{+}$.
  \item $ANN^{-}$ is equivalent to the LMNN, but the $ANN^{-}$ has achieved better performance of LMNN. That is because $ANN^{-}$ uses the average distance to determine whether the local neighborhood is pure. As a result, $ANN^{-}$ has a better searching space than that of LMNN.
  \item PNCA has achieved better result than the NCA in most of the data sets. That may be because the parameter $\alpha$ has the ability to adjust the mode to suit the distribution of data set better.
\end{itemize}
\subsection{Running Time of the Proposed Method}
\begin{figure}[t]
 \centering
\includegraphics[width=0.9\linewidth]{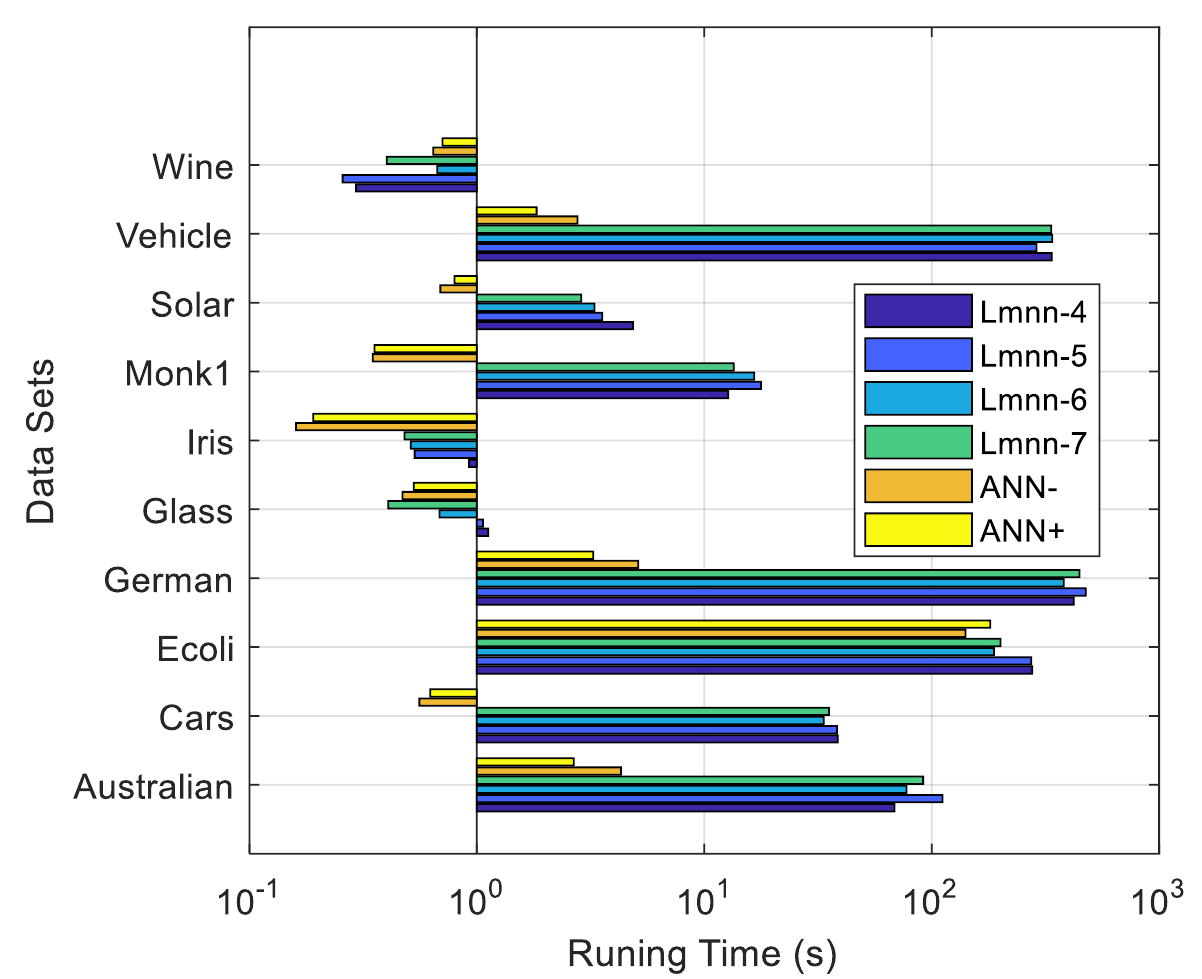}
\caption{The running times of different methods on different data sets.  The Lmnn-4,  Lmnn-5, Lmnn-6, Lmnn-7 represent LMNN with target neighbor number $\vert \mathcal{T}_i \vert= \{4,5,6,7\}$, respectively. $ANN^{+}$ and $ANN^{-}$ represent ANN with parameter $\alpha >0$ and $\alpha <0$, respectively.}
\label{fig5_runing}
\end{figure}

In this section, we compare the running times of the proposed method with LMNN. Since the solution procedures of $ANN^{+}$ and $ANN^{-}$ may have different iteration number, so we report the running times of them separately. For $ANN^{+}$ and $ANN^{-}$, all of the samples are input in the algorithms, that is to say both $\mathcal{S}_i$ and $\mathcal{D}_i$ of each inquiry sample $\textbf{x}_i$ are constructed to its largest volume. For LMNN, the number of target neighbors $\vert \mathcal{P}_i\vert$ is a essential factor for affecting the running time, we report the running time of LMNN with $\mathcal{P}_i = \{4,5,6,7\}$. The parameters of regularization term of $ANN^+$, $ANN^-$ and LMNN are set as $1/(N^2)$,  $1/(N^2)$ and $0.5$, respectively. The $\alpha$ and $\gamma$ in $ANN^{-}$ and $ANN^{+}$ are set as $1$ (or $-1$). For the three methods, each algorithm is performed 30 runs, and the average running time has recorded. The results are shown in Fig.\ref{fig5_runing}. As seen from the Fig. \ref{fig5_runing}, the $ANN^{-}$ and $ANN^{+}$ run much faster than the LMNN. In theoretic analysis, the running time of $ANN^{-}$ (or $ANN^{+}$) may be the $\frac{1}{\vert \mathcal{P}_i\vert}$ of that of LMNN, however, in practice we find that the proposed method is faster than LMNN by an order of magnitude. That is because we use the hinge loss function to penalize the constraint which involved in $N$ samples. When the hinge loss is not triggered, $N$ distance computations are removed. However, in LMNN one hinge loss only triggers $2$ distance computations. Anther possible reason is that the objective function of the proposed method is more smooth than that of LMNN, so the convergence speed of the proposed is faster than LMNN.\\
\indent Besides, we can observe that $ANN^{-}$ runs faster than $ANN^{+}$ in most of the data sets, but the difference is not very significantly. That may be because $ANN^{-}$ is a convex problem which is easier to find the convergence point than $ANN^{+}$ which is a non-convex one.\\
\begin{table*}[!htbp]
\renewcommand{\arraystretch}{1.7} 
\setlength{\tabcolsep}{2pt}  
\centering
\caption{Comparison of Different Methods on Several Data Sets.}
\label{ta1_result}
\footnotesize
\begin{tabular}{ccccccccc|ccc}
\toprule
&\multicolumn{8}{c|}{Baselines}&\multicolumn{3}{c}{Our Methods}\\
\midrule
Dataset &NCA&LMNN&ITML&LDML&SCML&RVML&GMML&BoostMetric&PNCA&$ANN^{+}$&$ANN^{-}$\\
\midrule
Australian&71.12$\pm$2.14&71.16$\pm$2.69&67.39$\pm$2.42&70.32$\pm$2.22&70.26$\pm$1.98&73.12$\pm$2.11&85.94$\pm$2.32&72.42$\pm$2.13&{79.15$\pm$2.42}&\textbf{83.68$\pm$2.35}&{80.84$\pm$2.27}\\
cars &80.96$\pm$2.21&83.33$\pm$1.94&81.38$\pm$2.12&80.16$\pm$2.47&83.11$\pm$2.28&82.68$\pm$2.51&84.91$\pm$2.26&84.16$\pm$2.17&{82.76$\pm$2.13}&\textbf{85.97$\pm$2.13}&{83.49$\pm$2.09}\\
Ecoli&76.32$\pm$1.59&79.15$\pm$1.43&81.52$\pm$1.27&81.22$\pm$1.79&80.17$\pm$1.93&81.32$\pm$1.45&76.22$\pm$1.24&77.47$\pm$1.63&81.54$\pm$1.72&{83.58$\pm$1.92}&\textbf{84.36$\pm$1.42}\\
German&67.23$\pm$2.41&78.51$\pm$2.21&74.51$\pm$2.65&77.31$\pm$2.49&75.22$\pm$2.37&74.14$\pm$2.55&71.62$\pm$2.04&77.11$\pm$2.29&{75.89$\pm$2.07}&\textbf{79.91$\pm$2.33}&{79.71$\pm$2.21}\\
Glass&70.09$\pm$1.34&71.43$\pm$1.56&65.88$\pm$1.37&71.11$\pm$1.29&66.32$\pm$1.54&71.21$\pm$1.57&62.61$\pm$1.61&70.22$\pm$1.76&{71.42$\pm$1.62}&{72.56$\pm$1.1.61}&\textbf{73.77$\pm$1.53}\\
Iris&95.76$\pm$1.89&96.11$\pm$1.96&96.67$\pm$2.01&96.21$\pm$2.16&95.22$\pm$2.32&96.11$\pm$2.22&97.47$\pm$2.18&96.23$\pm$2.04&{96.89$\pm$2.09}&\textbf{97.89$\pm$2.09}&{96.89$\pm$2.09}\\
Isolet&83.9$\pm$2.03&87.57$\pm$2.12&    84.05$\pm$2.10& 85.17$\pm$2.17&86.28$\pm$1.98&88.06$\pm$2.32&82.62$\pm$2.19&86.38$\pm$1.82&    90.31$\pm$2.24&\textbf{93.85$\pm$1.95}&{88.38$\pm$2.03}\\
Monk1&83.42$\pm$1.72&86.27$\pm$1.88&86.84$\pm$1.64&86.11$\pm$1.82&86.64$\pm$1.88&84.34$\pm$1.76&89.16$\pm$1.73&85.43$\pm$1.62&85.62$\pm$1.59&\textbf{91.74$\pm$1.74}&{87.43$\pm$1.74}\\
Solar&68.21$\pm$2.38&70.11$\pm$2.34&62.12$\pm$2.41&65.12$\pm$2.25&65.22$\pm$2.18&66.33$\pm$2.49&64.05$\pm$2.53&63.61$\pm$2.41&71.52$\pm$2.21&{69.57$\pm$2.25}&\textbf{96.99$\pm$2.50}\\
Vehicle&71.22$\pm$2.22&73.96$\pm$2.18&68.79$\pm$2.19&72.21$\pm$2.26&72.91$\pm$2.52&70.12$\pm$2.24&\textbf{78.15$\pm$2.17}&72.18$\pm$2.19&72.49$\pm$2.15&{76.78$\pm$2.31}&{75.79$\pm$2.24}\\
Wine&87.63$\pm$2.24&89.14$\pm$2.12&89.32$\pm$2.13&89.41$\pm$2.32&88.34$\pm$2.39&90.12$\pm$2.11&86.32$\pm$2.22&90.13$\pm$2.32&92.27 $\pm$2.12&{98.15$\pm$2.13}&\textbf{98.15$\pm$2.19}\\
Pendigits&94.12$\pm$1.36&97.72$\pm$1.46&94.24$\pm$1.46&95.32$\pm$1.45&96.43$\pm$1.52&98.22$\pm$1.62&94.21$\pm$1.05&96.53$\pm$2.11&95.46$\pm$2.21&98.43$\pm$1.62&\textbf{98.54$\pm$1.29}\\
Coil20&94.32$\pm$2.01&95.21$\pm$2.12&94.17$\pm$2.16&93.25$\pm$2.21&\textbf{96.81$\pm$2.09}&96.52$\pm$1.96&93.42$\pm$2.21&93.26$\pm$1.95&94.72$\pm$2.15&96.47$\pm$2.18&96.51$\pm$2.03\\
Letter&93.74$\pm$2.41&95.63$\pm$2.47&93.83$\pm$2.44&94.42$\pm$2.21&93.31$\pm$2.31&95.72$\pm$3.61&94.12$\pm$2.68&95.11$\pm$2.45&94.26$\pm$2.31&{95.15$\pm$2.27}&\textbf{95.49$\pm$2.27}\\
USPS&92.45$\pm$2.21&94.52$\pm$2.31&91.07$\pm$2.12&92.14$\pm$2.05&93.51$\pm$2.26&92.72$\pm$2.62&94.32$\pm$2.31&95.24$\pm$2.11&93.57$\pm$2.15&95.21$\pm$2.06&\textbf{95.54$\pm$2.02}\\
\bottomrule
\end{tabular}
\end{table*}
\section{Deep Feature Extraction}
\indent In this section, we would employ the proposed formulation to extract the deep features for many tasks in computer version.\\
\indent Since our work give a general formulation for triplet-based distance metric learning, when we replace the Mahalanobis distance $d^2_\textbf{M}(\textbf{x}_i,\textbf{x}_j)$ by the Euclidean distance between two deep features, i.e., $\vert f(\textbf{x}_i) -f(\textbf{x}_j)\vert^2$, we can obtain a deep metric learning model. Therefore, we employ this model to extract deep features.\\
\section{Conclusion}
In this paper, we propose a novel distance metric learning model named adaptive nearest neighbor (ANN). This new distance metric learning method has a very interesting property. That is by setting different parameter value, the proposed method can be seen as the improvement of the existing state-of-the-art methods, including the LMNN and NCA. Compared with the original version of LMNN, the proposed method has a boarder searching space for the appropriate metric matrix. Compared with NCA, the proposed method can match the distribution of training data more accurately. We evaluated our algorithm on several data sets with various sizes and difficulties. Compared with the state-of-the-art methods, the proposed methods have achieved better classification results on most of data sets. Compare with the LMNN, our method performs much faster.\\

\bibliographystyle{ieeetr}
\bibliography{bare_jrnl}

\ifCLASSOPTIONcaptionsoff
  \newpage
\fi

\begin{IEEEbiography}{Kun Song}
received his master degree from the Northwestern Polytechnical University, Xi'an, China, in 2015. He is currently pursuing the Ph.D degree at Northwestern Polytechnical University. His research interests are computer vision and machine learning.
\end{IEEEbiography}

\begin{IEEEbiography}{Junwei Han}
(M'12 - SM'15) received the Ph.D degree in pattern recognition and intelligent systems from the School of Automation, Northwestern Polytechnical University in 2003. He is a currently a Professor with Northwestern Polytechnical University, Xi'an, China. His research interests include multimedia processing and brain imaging analysis. He is an Associate Editor of the IEEE Transactions on Human-Machine Systems, Neurocomputing, Machine Vision and Applications, and Multidimensional Systems and Signal Processing.
\end{IEEEbiography}

\end{document}